\documentclass[letterpaper,10pt,twocolumn]{article}
\usepackage[T1]{fontenc}
\usepackage[utf8]{inputenc}
\usepackage[english]{babel}
\usepackage{lmodern}
\usepackage[letterpaper,margin=0.72in,columnsep=0.24in]{geometry}
\usepackage{amsmath,amssymb}
\usepackage{algorithm}
\usepackage{algpseudocode}
\usepackage{cite}
\usepackage{float}
\usepackage{microtype}
\usepackage{booktabs,array}
\usepackage{graphicx}
\usepackage[font=small,labelfont=bf]{caption}
\IfFileExists{xurl.sty}{\usepackage{xurl}}{\usepackage{url}}
\usepackage[unicode,hidelinks]{hyperref}
\usepackage{bookmark}
\graphicspath{{figures/}}
\hypersetup{
  pdftitle={Making Self-Driving Laboratories Auditable: A Unified Framework for Hybrid Actions, Measurements, and Safe Execution},
  pdfsubject={English preprint: system design and engineering report},
  pdfauthor={Zihan Zhang, Zeyu Wu, Xin Zhang, Mingyu Luo, Haohui Que, Tong Zhu},
  pdfcreator={LaTeX}
}
\providecommand{\tightlist}{\setlength{\itemsep}{0pt}\setlength{\parskip}{0pt}}
\author{%
  \begin{tabular}{c}
  \normalsize Zihan Zhang$^{1,2,5}$\quad Zeyu Wu$^{5}$\quad Xin Zhang$^{1,4,5}$\\[3pt]
  \normalsize Mingyu Luo$^{5}$\quad Haohui Que$^{1,3,5,*}$\quad Tong Zhu$^{1,3,5,*}$\\[6pt]
  \small $^{1}$SII--ECNU Joint Autonomous Chemistry Laboratory, Shanghai Innovation Institute\\
  \small $^{2}$Nankai University\quad $^{3}$East China Normal University\\
  \small $^{4}$Shanghai Jiao Tong University\quad $^{5}$Shanghai Lixun Intelligent Technology Co.,Ltd\\[4pt]
  \small Haohui Que: \texttt{harveyque@outlook.com}\quad
  Tong Zhu: \texttt{tongzhu.work@gmail.com}
  \end{tabular}%
}
\date{}

\begin{document}
\title{\bfseries Making Self-Driving Laboratories Auditable:\\
A Unified Framework for Hybrid Actions,\\
Measurements, and Safe Execution}
\twocolumn[
\maketitle
\begin{abstract}
Self-driving laboratories (SDLs) are laboratory systems that couple experimental planning, automated execution, and measurement feedback in a closed loop. Integrating flexible robot control with the typed application programming interfaces (APIs) and continuous measurements exposed by laboratory instruments remains an execution-interface challenge. To address this challenge, we propose a traceable framework comprising four components: LAM (Lab Action Model), OpenSDL, Safe-SDL, and L3-Bench. LAM is a proposed measurement-conditioned policy architecture that extends vision-language-action paradigms to generate both continuous robot trajectories and typed instrument API calls. OpenSDL is an execution substrate with an initial control implementation, designed to map role-level operations to heterogeneous devices, coordinate execution, and record timestamped causal events. The proposed Safe-SDL runtime gate combines pre-dispatch schema and resource checks with signal temporal logic monitoring during execution, including extended waiting periods. L3-Bench (Level-3 Laboratory Benchmark) specifies an episode-level evaluation protocol covering physical outcomes, timing, reproducibility, safety violations, and fault recovery. Together, these components provide an end-to-end design for linking hybrid action generation, monitored device execution, measurement feedback, and experimental evaluation, addressing coordination and traceability challenges in SDL workflows.

\end{abstract}
\vspace{1.0em}
]
\begingroup
\renewcommand{\thefootnote}{\fnsymbol{footnote}}
\footnotetext[1]{Corresponding authors.}
\endgroup
\section{Introduction}\label{introduction}

Self-driving laboratories (SDLs) are laboratory systems that couple experimental intent and planning to automated execution and measurement feedback in a closed loop~\cite{abolhasani2023}. This approach can accelerate materials discovery and reaction optimization while reducing the need for repeated manual intervention~\cite{abolhasani2023,slattery2024,szymanski2023}. Their defining operation is a repeated decision cycle: (i) a machine-readable objective is translated into experimental operations; (ii) those operations are executed by instruments and robots; (iii) measurements and device states are collected; and (iv) the resulting evidence is used to inform the next decision. Through this cycle, experimental throughput can be increased and condition search or hypothesis testing conducted systematically across many trials~\cite{king2004,steiner2019,slattery2024,szymanski2023}. To implement this cycle, practical SDLs combine planning tools for step (i), device control for step (ii), sensing and data acquisition for step (iii), and data analysis and decision-making for step (iv). Data and provenance management link these stages, while human supervision supports oversight and intervention throughout the cycle~\cite{abolhasani2023}.

Representative developments have showcased how experimental procedures and robot actions can be represented, executed, and recorded. In 2019, Steiner et al. described Chemputer, which encoded chemical procedures as executable programs, separating hardware-independent procedures from platform-specific connectivity and device drivers~\cite{steiner2019}. In 2023, Boiko et al. introduced Coscientist, which combined language-model reasoning with documentation search, code execution, and laboratory APIs to plan and perform chemical experiments~\cite{boiko2023}. In 2024, Kim et al. introduced OpenVLA, a vision-language-action (VLA) model connecting visual observations and language instructions to learned robot control~\cite{kim2024}. In the same year, Bai et al. developed a dynamic knowledge graph framework for distributed SDLs that supports shared ontologies and cycle-level provenance, with limited granularity for exact operation timing under hardware constraints~\cite{bai2024}. In 2026, Ren et al. introduced LabVLA, which applied VLA policies to laboratory protocols while acknowledging that measurement-driven protocol revision had not yet been demonstrated~\cite{ren2026}. IvoryOS provides an interoperable web interface for orchestrating Python-based SDLs~\cite{zhang2025}. Established standards address device connectivity and provenance. SiLA 2 (Standardization in Lab Automation) describes instrument capabilities through typed Features, Commands, and Properties, supporting discovery, command execution, and property subscriptions~\cite{juchli2022}. Provenance standards provide a formal basis for representing auditable records. The World Wide Web Consortium's PROV Ontology (W3C PROV-O) represents entities, activities, and agents through generation, usage, attribution, and qualified relations carrying roles and time information~\cite{lebo2013}. PROV-CONSTRAINTS defines validity, equivalence, and event-ordering constraints for provenance records~\cite{cheney2013constraints}; the companion PROV-SEM Working Group Note provides first-order semantics for the PROV data model~\cite{cheney2013semantics}. ProvONE extends PROV with scientific workflow structure and execution provenance~\cite{cao2016provone}, while Research Object Crate (RO-Crate) packages research artefacts with Schema.org-based structured metadata~\cite{soilandreyes2022rocrate}.

These approaches establish provenance representation and consistency rules. Our execution-interface contribution concerns their coupling to measurement-conditioned hybrid actions, pre-dispatch gating, clock-domain-aware timestamping, and timed safety monitoring. We position SDL-Traj as a proposed PROV-O-aligned laboratory profile, reusing core provenance relations and adding application-specific terms for device capabilities, monotonic-clock metadata, and signal temporal logic (STL) robustness. Explicit vocabulary mappings, PROV-O export, SiLA 2 adapters, and conformance tests remain implementation targets. For example, reaction and cooling waits require monitoring to remain active and link fresh or stale measurements, pending calls, verified completion events, and gate decisions. Such execution records must preserve both scientific context and the conditions under which physical actions were admitted. These requirements motivate a shared interface linking measurement-conditioned action generation with execution, monitoring, and evaluation. We therefore propose a traceable framework comprising LAM (Lab Action Model), OpenSDL, Safe-SDL, and L3-Bench.

LAM is designed to generate a hybrid output consisting of continuous robot trajectories and typed instrument API calls from task instructions, visual observations, robot states, and instrument measurements. Its proposed architecture extends the VLA framework with typed telemetry and shares multimodal context between a continuous motion expert and a discrete function-calling head. OpenSDL, with an initial control implementation, is designed to map role-level operations to heterogeneous devices and coordinate operations according to their dependencies, waiting requirements, and completion conditions. Its event model links measurements, device states, and execution outcomes to dispatched commands and runtime decisions in timestamped causal records. In the proposed loop, updated observations and confirmed events provide context for subsequent LAM action proposals. Safe-SDL is the proposed runtime gate for schema and resource checks before dispatch and monitoring of formalized temporal constraints throughout execution, including extended waiting periods. Timeouts, degraded measurement quality, or insufficient safety margins are intended to trigger action blocking and predefined degradation or safe-abort procedures. L3-Bench (Level-3 Laboratory Benchmark) specifies an episode-level evaluation protocol covering protocol-specific physical outcomes, timing, reproducibility, safety violations, and fault recovery. Here, L3 denotes measurement-conditioned adaptation within an experiment, as defined in the task-specific classification of Section~\ref{autonomy-levels}. End-to-end LAM training, cross-device transfer experiments, and systematic hardware and safety validation remain to be completed; improvements in throughput, adaptability, and fault recovery are hypotheses to be tested through the proposed evaluation.

\section{Problem Formulation and Control Interface}\label{problem-formulation-and-system-philosophy}

\subsection{Autonomy Levels}\label{autonomy-levels}

We adopt a task-specific working L1--L4 autonomy classification that complements published SDL autonomy taxonomies~\cite{tobias2025,hung2024}. These levels classify execution capability; SDLs additionally require the planning-and-feedback loop defined above. L1--L4 describe increasing adaptation: L1 denotes scripted execution or teleoperation, L2 visual protocol execution, L3 measurement-conditioned adaptation, and L4 hypothesis optimization across batches with multi-instrument coordination. We focus on implementable and measurable \textbf{L3} behavior in the near term, leaving L4 for future work.

\subsection{Hybrid State--Action and Event Semantics}\label{hybrid-stateaction-space}

We represent the laboratory process as a partially observable Markov decision process (POMDP), a standard model for sequential decision-making when the physical state cannot be observed completely:
\[
\mathcal{M}=\langle\mathcal{S},\mathcal{A},\mathcal{T},\mathcal{R},\Omega,\mathcal{O},\gamma\rangle .
\]
Here, $\mathcal{S}$ is the latent physical state of the laboratory, $\mathcal{A}$ is the hybrid candidate-action space, $\mathcal{T}(s'\mid s,a)$ is the transition model describing how action $a$ changes state $s$ to $s'$, and $\mathcal{R}(s,a)$ is the task or operational reward. $\Omega$ is the observation space, $\mathcal{O}(o\mid s',a)$ is the observation model, and $\gamma\in[0,1]$ discounts delayed reward. The notation captures temperatures, concentrations, device queues, and completion states that are partially observed and may be delayed or noisy; the current implementation leaves the transition and reward models unspecified.
At time $t$, the observation is $o_t=\{I_t,q_t,M_t\}$, where $I_t$ contains synchronized visual frames, $q_t$ is the proprioceptive state, and $M_t$ contains measurement streams from heterogeneous instruments. The action space is the Cartesian product
\[
\mathcal{A}=\mathcal{A}_{\text{arm}}\times\mathcal{A}_{\text{inst}} .
\]
The arm-action space includes a no-motion option, and the instrument-action space contains finite, possibly empty collections of calls. Feasible combinations depend on resource and safety constraints. A continuous action $A_{\text{arm}}\in\mathbb{R}^{H_a\times D_a}$ contains $H_a$ control steps and $D_a$ trajectory dimensions, with recorded coordinate frame, control period, and calibration version. An individual instrument call $A_{\text{inst}}=\langle\text{Role},\text{Op},\Theta\rangle$ is a typed instrument call: \texttt{Role} identifies the capability, \texttt{Op} identifies the operation such as \texttt{Set}, \texttt{Execute}, or \texttt{Read}, and $\Theta$ contains validated parameters and units. The two action types share a time base and event ledger. Empty outputs allow monitoring during waiting periods.

To give this Cartesian product executable semantics, the output of a decision cycle is written as
\[
a_t=\left(\hat A_{\mathrm{arm},t:t+H_a-1},\mathcal{C}_t\right),\qquad
\mathcal{C}_t=\{c_i\}_{i=1}^{N_t},
\]
where each instrument call includes a role, an operation, parameters, a waiting policy, and an idempotency key:
\[
c_i=\langle r_i,o_i,\theta_i,\delta_i,\kappa_i\rangle .
\]
Here, $r_i$ is a semantic capability in the role dictionary, $o_i\in\{\texttt{Set},\texttt{Execute},\texttt{Read}\}$ is the requested operation, $\theta_i$ contains typed parameters, $\delta_i$ specifies timeout, retry, and completion conditions, and $\kappa_i$ supports deduplication and transaction association. The action compiler checks units, ranges, resource locks, and preconditions before mapping a role call to a concrete DeviceProvider (an adapter that implements a role for one physical or simulated device). Role-preserving substitution retains high-level action labels; device calibration, latency, and physical limits still require compatibility tests.

States, actions, and events share a common timeline. Observations, candidate actions, and device returns are recorded as events $e_j$, linked by episode IDs, timestamps, and causal parent events. The proposed ledger combines device timestamps, a runtime monotonic clock, and causal event IDs. Clock offsets and drift are recorded explicitly; receive order alone cannot establish cross-device causality. In this paper, an event is therefore the auditable record of one observation, proposal, gate decision, dispatch, return, or compensation step recorded in the event ledger. This design distinguishes an incorrect model proposal, a safety-blocked action, and a device execution failure, and allows replay to reconstruct the causal sequence from proposal through approval to completion.

\subsection{Design Requirements}\label{design-requirements}

We express the laboratory control interface as five checkable invariants. Semantic portability requires policies to depend on roles, capabilities, and units. Action composability requires continuous trajectories and discrete calls to carry preconditions, resource requirements, and completion signals so that they can form a directed acyclic graph (DAG) execution plan. Temporal consistency requires visual frames, proprioception, telemetry, and action events to share monotonic timestamps with explicit measurement delays and waiting intervals. Safety gating requires every action to pass through the same observable safety kernel before submission to the bus; device interlocks govern actions independently of model confidence. Replayability requires raw inputs, compiled device commands, return bytes, and safety decisions to be linked to the same episode for offline replay and difference analysis.

These invariants define an auditable protocol contribution and establish module boundaries. LAM generates candidate actions from context, while OpenSDL checks their semantic, temporal, and safety properties; the modules exchange data only through a versioned schema. If an invariant depends on manual action, that intervention must be reported in evaluation.

The evaluation addresses four research questions (RQs):
\begin{list}{\normalfont RQ\arabic{enumi}.}{%
\usecounter{enumi}%
\renewcommand{\theenumi}{RQ\arabic{enumi}}%
\setlength{\leftmargin}{2.8em}%
\setlength{\labelwidth}{2.4em}%
\setlength{\labelsep}{0.4em}%
\setlength{\itemindent}{0pt}%
\setlength{\topsep}{3pt}%
\setlength{\partopsep}{0pt}%
\setlength{\itemsep}{0pt}%
\setlength{\parsep}{0pt}%
}
\item\label{rq:measurement} Does measurement-token conditioning improve adaptation to measurement shifts?
\item\label{rq:coordination} Does concurrent hybrid action dispatch reduce completion time while respecting resource constraints?
\item\label{rq:safety} Does runtime gating contain faults and support recovery at a measurable throughput cost?
\item\label{rq:portability} Can role-preserving device substitution maintain action semantics, with fixed-manifest replay reproducing execution histories and metrics?
\end{list}

\section{LAM: Lab Action Model}\label{lam-lab-action-model}

\begin{figure*}[t]
\centering
\includegraphics[width=\textwidth]{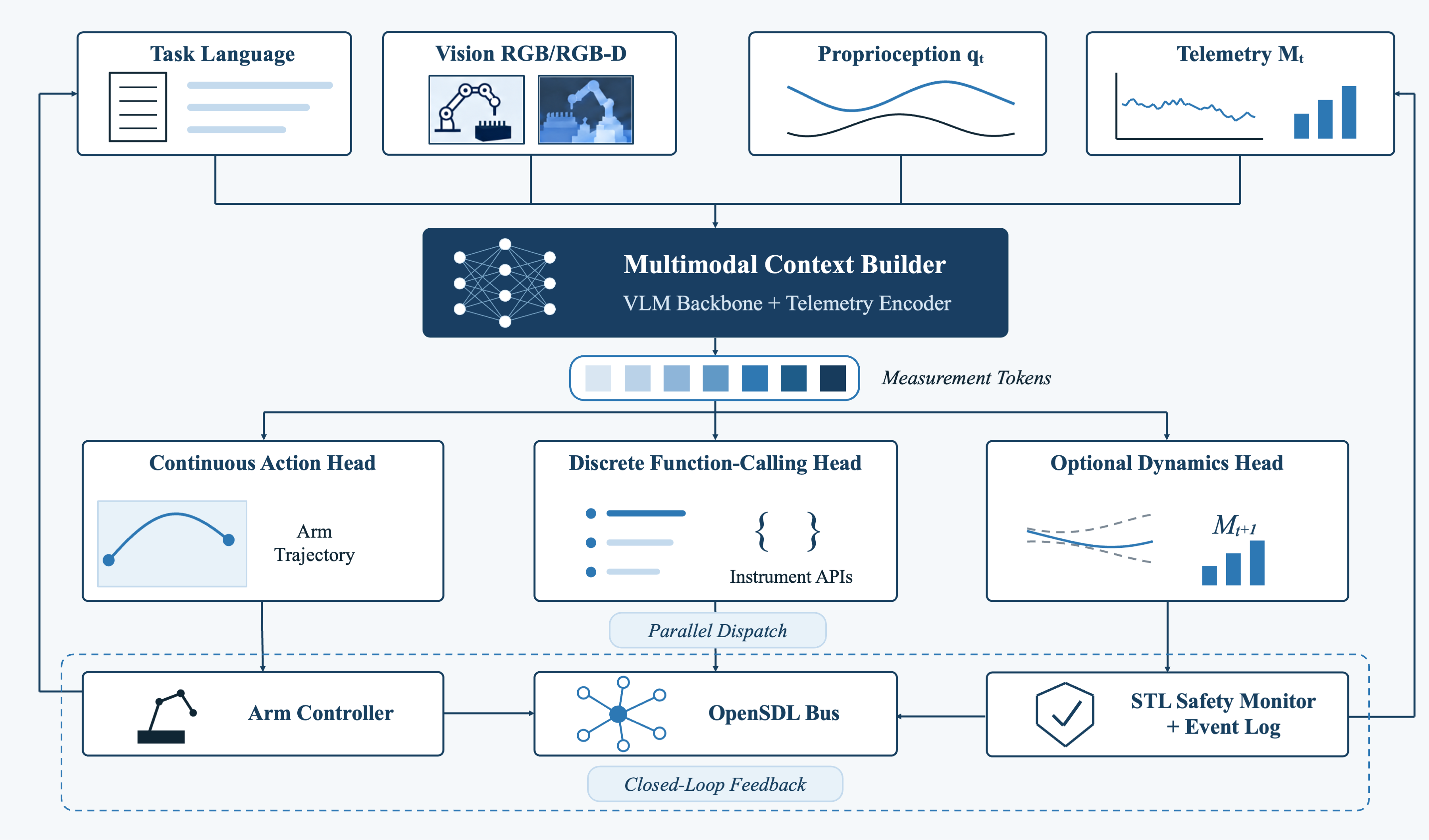}
\caption{LAM architecture for measurement-conditioned hybrid laboratory control. Task language, RGB/RGB-D observations, robot proprioception $q_t$, and time-stamped instrument telemetry $M_t$ enter a multimodal context builder comprising a vision-language-model (VLM) backbone and a telemetry encoder. Typed measurement tokens retain device roles, units, sampling times, and data-quality information within the shared context. A continuous action head generates robot-trajectory chunks, a discrete function-calling head generates typed instrument application programming interface (API) calls, and an optional dynamics head predicts the next measurement state $M_{t+1}$. The execution layer connects the arm controller and the signal temporal logic (STL) safety monitor with event logging to the OpenSDL bus. Parallel dispatch coordinates compatible actions, while updated observations and confirmed execution events supply feedback for subsequent decisions.}
\label{fig:lam-architecture}
\end{figure*}

\subsection{Model Structure and Multimodal Context Construction}\label{model-structure}

LAM (Lab Action Model) is the policy interface proposed in this paper. The proposed LAM architecture comprises three cooperating modules:

\begin{enumerate}
\def\labelenumi{\arabic{enumi}.}
\tightlist
\item \textbf{Multimodal context backbone (target size: approximately 3B parameters; VLM denotes vision-language model).}
Encodes task language, visual context, trajectory history, and device states.
\item \textbf{Continuous motion expert (target size: approximately 300M parameters).}
Uses flow matching to generate trajectory chunks, following the VLA motion-modeling approach exemplified by $\pi_0$~\cite{black2024}; obstacle avoidance is checked by the execution layer.
\item \textbf{Discrete function-calling head.}
Generates standardized \texttt{Role--Op--Parameters} calls with device waiting, failure, and retry semantics.
\end{enumerate}

At each decision cycle, the backbone receives a fixed-length history window and the current task constraints. The two action heads share context and use separate output vocabularies. The motion expert generates a trajectory chunk for the next $H_a$ steps, while the function-calling head generates zero or more calls with resource declarations. The scheduler determines which actions may execute concurrently from device occupancy and action dependencies. This decomposition retains the VLA's language and visual representations and keeps instrument calls in their typed API representation.

Figure~\ref{fig:lam-architecture} shows the computation from heterogeneous observations to concurrent actions.

The shared backbone first assembles asynchronous observations into a unified context. Let the language, vision, proprioception, and telemetry encoders produce $X^L_t,X^V_t,X^q_t,X^M_t$, respectively. Each token carries modality, role, unit, sampling-time, and quality information. Missing or stale entries receive explicit status embeddings and restricted attention access:
\[
X_t=\operatorname{Concat}\left(X^L_t,X^V_t,X^q_t,X^M_t,X^{\mathrm{hist}}_t\right).
\]
Here, $X^{\mathrm{hist}}_t$ stores the most recent $w$ confirmed events; $\operatorname{Concat}$ preserves their modality order while forming one attention sequence. The superscripts $L,V,q,M$ identify language, vision, proprioception, and measurement tokens, respectively. For missing or stale sensors, the encoder emits explicit missing/stale flags; interpolated values carry the same status. This convention distinguishes unavailable measurements from measurements ignored by the model during cross-device transfer.

\subsection{Hybrid Action Decoding and Uncertainty Handling}\label{hybrid-action-decoding}

Given task specification $\mathcal{G}$ and the shared representation $h_t=\operatorname{VLM}_\theta(X_t,\mathcal{G})$, the continuous and discrete heads compute
\[
\hat A_{\mathrm{arm}}=f_{\mathrm{flow}}(h_t, q_t),\qquad
\hat{\mathcal{C}}_t=f_{\mathrm{call}}(h_t,\mathcal{R}_t),
\]
where $\mathcal{R}_t$ is the set of currently available roles and capabilities, $f_{\mathrm{flow}}$ maps context and robot state to a continuous trajectory, and $f_{\mathrm{call}}$ maps context and available roles to typed calls. The parameter $\theta$ denotes trainable model weights. The function-calling head uses schema-constrained structured decoding: roles, operations, and parameters must pass type checks before entering the execution queue. The scheduler inserts explicit barriers for calls with resource conflicts, while read-only measurements and workspace-independent actions may overlap with robot trajectories. The final action is an executable subgraph obtained by resolving dependencies between the two heads.

Laboratory policies face risks from out-of-distribution observations and sensor distortion as well as average prediction error. The proposed confidence interface reports an action-level score $u_t$ and device-quality summary $g_t$. Discrete-call token scores and continuous-action uncertainty require head-specific estimators, calibrated against held-out action failures. Estimator construction, calibration error, and abstention thresholds must accompany a trained model. At runtime, confidence below a pre-registered threshold or stale telemetry indicated by $g_t$ triggers read-only operation, a request for human confirmation, or safe degradation. Rejected proposals are logged as abstentions and reported separately from device retries. Their episodes remain in episode-level accounting, including termination outcomes and human interventions.

\subsection{Measurement-Conditioned Training Objective}\label{measurement-token-injection}

The Telemetry Encoder projects calibrated measurements into the shared context. A scale-preserving specification is
\[
e_M=\mathrm{MLP}(\mathrm{Scale}_{\kappa}(M_t))+E_{\mathrm{meta}}
\in\mathbb{R}^{K\times D_{\mathrm{model}}}.
\]
Here $K$ counts measurement tokens, $D_{\mathrm{model}}$ is the embedding dimension, and $\kappa$ fixes unit conversions and channel scales from training data or calibration. $E_{\mathrm{meta}}$ embeds roles, units, timestamps, and quality flags. Raw values and scaling metadata remain available to the safety monitor. Fixed scaling retains absolute magnitudes at the encoder input for threshold-dependent decisions. The resulting $e_M$ supplies $X^M_t$. The optional dynamics head predicts $M_{t+1}$ at a fixed, recorded forecast offset, using valid-channel masks.

For normalized action coordinates, a concrete conditional flow-matching objective follows the linear interpolation used in flow-based VLA training~\cite{black2024}:
\[
\begin{aligned}
Z_u&=(1-u)\xi+uA^*_{\mathrm{arm}},\\
\mathcal{L}_{\mathrm{flow}}&=\mathbb{E}\!\left[\left\|
v_\theta(Z_u,u\mid h_t,q_t)
-(A^*_{\mathrm{arm}}-\xi)\right\|_F^2\right].
\end{aligned}
\]
The expectation samples demonstrated chunks, standard Gaussian noise $\xi$ of the same shape, and flow time $u\sim\mathcal{U}(0,1)$. Flow time is separate from physical time $t$. The vector field $v_\theta$ predicts the interpolation velocity; integration from noise at $u=0$ to $u=1$ produces a chunk that is mapped back to physical coordinates. The Frobenius norm sums squared errors over control steps and dimensions. This specifies a training objective; model training remains pending.

The hybrid training loss is
\[
\begin{aligned}
\mathcal{L}_{\mathrm{total}}={}&\mathcal{L}_{\mathrm{flow}}
+\alpha\mathcal{L}_{\mathrm{CE}}(\mathcal{C}^*_t,p_\theta)\\
&+\beta\mathcal{L}_{\mathrm{pred}}(M_{t+1},\hat M_{t+1}).
\end{aligned}
\]
Here $\mathcal{C}^*_t$ contains validated target calls, and cross-entropy supervises their serialized tokens using conditional probabilities $p_\theta$ before decoding. The optional prediction loss is mean squared error on available, fixed-scale measurement targets and is skipped when none are valid; $\alpha,\beta\ge0$ weight the terms. Training records retain observations, actions, device returns, waiting intervals, exception labels, and safety decisions.

Cross-modal consistency and safety-supervised auxiliary losses remain possible extensions. Their label construction, differentiable definitions, and ablations must be specified before inclusion in a trained model. Runtime gating remains an independent execution requirement.

\subsection{Data Curation and Splits}\label{data-curation-and-splits}

Training, calibration, and test partitions group entire experimental batches, including their episodes and adjacent frames. Normalization statistics are fitted on training data. Device-transfer tests hold out the target provider during training. Each episode retains task descriptions, device and calibration metadata, observations available by decision time, candidate actions, device returns, waiting intervals, safety decisions, and termination reasons. Four evaluation settings cover nominal operation, measurement shift (\ref{rq:measurement}), device transfer with preserved role capabilities and a substituted DeviceProvider (\ref{rq:portability}), and fault injection (\ref{rq:safety}). Evaluation conditions share a schema version; overlapping shift conditions are flagged and reported separately. The concurrency ablation for \ref{rq:coordination} uses matched task instances and seeds, with fresh rollouts for each policy.

\section{OpenSDL: Unified Infrastructure and Runtime
Safety}\label{opensdl-unified-infrastructure-and-runtime-safety}

\subsection{Unified Research
Infrastructure}\label{unified-research-infrastructure}

The initial OpenSDL implementation provides a Rust control engine, protocol adapters, MQTT/serial/TCP transports, ESP32 bridging, and a SQLite store for commands, device events, and serial bytes. The proposed research interface extends this substrate with SDL-Traj trajectories aligned to LeRobot-style data~\cite{cadene2026}, semantic role/DeviceProvider contracts, and causal safety records. The Mother Node coordinates device routing; the target event model links observations and decisions across simulation, hardware-in-the-loop (HIL) evaluation, and hardware. These research extensions and their conformance tests remain distinct from the implemented control substrate.

\subsection{Role, DeviceProvider, and Event Ledger}\label{role-and-deviceprovider-contract}

The role abstraction expresses physical device capabilities through a stable semantic interface. A role descriptor declares at least its name, supported operations, parameter schema, units, and safety ranges. A DeviceProvider additionally declares protocol versions, connection status, available resources, and rollback capabilities. A call is compiled only when both the descriptor and provider accept it. For example, a \texttt{Set} call on the \texttt{heater} role specifies target temperature and hold time; serial registers or vendor software development kit (SDK) calls remain internal to the provider. Two providers implementing the same role produce consistent SDL-Traj action labels, while calibration coefficients and device diagnostics are stored as metadata.

The contract distinguishes three return states: \texttt{accepted} means that a command has entered the device queue, \texttt{completed} means that its completion condition has been satisfied, and \texttt{failed} means that the device or communication layer cannot complete it. A \texttt{Read} call may return a stream or a snapshot and must include units, sampling time, quality flags, and calibration version. Operations with irreversible physical effects must be marked \texttt{non-compensable} in the descriptor, allowing the runtime to require a higher approval level or prohibit automatic retries.

The implemented store uses wall-clock ingestion timestamps. The proposed causal ledger adds clock calibration, episode links, and six lifecycle event types: proposal, gate, dispatch, acknowledge, complete, and compensate. Each event contains a minimal traceability tuple:
\[
\begin{aligned}
e_j=\langle&\mathrm{episode},\mathrm{event},\mathrm{type},t_j^{\mathrm{mono}},t_j^{\mathrm{device}},\\
&\mathrm{role},\mathrm{op},\mathrm{args},\mathrm{status},\mathrm{parents}\rangle .
\end{aligned}
\]
Here, \texttt{episode} identifies one complete task attempt, \texttt{event} is the unique event ID, and \texttt{type} identifies one of the six lifecycle stages. The timestamps $t_j^{\mathrm{mono}}$ and $t_j^{\mathrm{device}}$ record runtime monotonic time and device-reported time; \texttt{role}, \texttt{op}, and \texttt{status} identify capability, operation, and outcome; \texttt{parents} records causal predecessor IDs, including all prerequisites at a join. \texttt{args} stores a summary of schema-validated parameters, while raw transmitted and received bytes are linked through content hashes. Safety-kernel inputs, robustness scores, and gating reasons are linked to the corresponding action's event ID, allowing an exception to be traced to specific observations and model versions.

The proposed replay protocol reconstructs causal event order in three modes: \textit{dry-run} performs only schema and safety checks; \textit{sim-replay} sends actions to the digital twin; and \textit{hardware-replay} permits only explicitly allowlisted operations approved by a human. Replay compares event sequences, state transitions, and measurement trajectories. Differences due to random sampling, clock drift, or device noise must be explicitly categorized in the report.

\begin{table*}[t]
\centering
\small
\setlength{\tabcolsep}{4pt}
\caption{Minimum fields required for an auditable SDL-Traj event.}
\label{tab:event-schema}
\begin{tabular}{p{0.18\textwidth}p{0.25\textwidth}p{0.45\textwidth}}
\toprule
\textbf{Field group} & \textbf{Examples} & \textbf{Audit invariant} \\
\midrule
Identity & episode ID, event ID, parent IDs & Every action and return is traceable to one episode and its causal predecessors. \\
Observation & RGB/RGB-D, $q_t$, $M_t$, quality flags & Sensors carry units, calibration version, sample time, and stale/missing status. \\
Action & role, op, typed parameters, idempotency key & The submitted action is identical to the schema-validated action used for evaluation. \\
Execution & provider, protocol version, ack, completion, raw bytes & Device-level failures are separated from model errors and safety blocks. \\
Safety & formula version, robustness, gate decision, operator override & A blocked or approved action has an explicit, replayable reason. \\
Environment & asset commit, configuration hash, seed, clock offset & A replay can reconstruct the software, asset, and timing context. \\
\bottomrule
\end{tabular}
\end{table*}

\subsection{Temporal Coordination and Runtime State Machine}\label{temporal-coordination-and-concurrency}

The scheduling unit is an action graph with resource and temporal constraints. Nodes declare device resources to read or write, expected duration, prerequisite events, and cancellation semantics; directed edges encode dependencies and barriers. Resource conflicts are resolved through reservations and an acyclic ordering before topological dispatch. Two writes to the same instrument use mutual exclusion; reads use snapshots or stream subscriptions; and workspace reservations coordinate robot access; independent motion checks and device interlocks enforce physical limits.

To prevent an uninterpretable intermediate state when one action succeeds while another times out, the proposed scheduler uses a logical prepare-and-dispatch protocol. Providers that support preparation acknowledge safety-approved proposals before execution is committed once prerequisite events are satisfied. This protocol coordinates software state; physical reversibility depends on the device and operation. If a node fails, the scheduler selects retry, compensate, or abort according to provider compensation actions and checkpoint policies. This mechanism accounts for potentially irreversible physical operations; such operations remain explicit in evaluation and human approval.

The proposed runtime state machine defines seven states with the following transition rules:
\begin{itemize}
\item \textsc{Observed} admits observations only after synchronization and data-quality checks.
\item \textsc{Proposed} permits LAM to generate candidate actions.
\item \textsc{Gated} performs schema, resource, and STL checks.
\item \textsc{Dispatched} waits for device acknowledgment, while \textsc{Verified} records completion and updates the checkpoint.
\item Timeouts, invalid measurements, or insufficient robustness trigger recovery or \textsc{SafeAbort}; \textsc{Compensated} is recorded only after recovery is verified.
\end{itemize}
Each transition records its triggering event and operator overrides, making the state machine itself an object of evaluation.

\subsection{Safe-SDL Runtime Gating}\label{safe-sdl-runtime-gating}

We impose an STL specification on continuous state trajectories:
\[
\begin{aligned}
\varphi_{\mathrm{safe}}=\Box_{[0,\tau]}\bigl[&
  (T\le T_{\mathrm{crit}}) \\
  &\wedge(\lVert F_{\mathrm{contact}}\rVert\le F_{\max}) \\
  &\wedge(d_{\mathrm{vial,obs}}\ge d_{\mathrm{safe}})\bigr].
\end{aligned}
\]
Here $T$ is measured temperature with limit $T_{\mathrm{crit}}$, $F_{\mathrm{contact}}$ is contact force with limit $F_{\max}$, and $d_{\mathrm{vial,obs}}$ is robot clearance from protected labware with minimum $d_{\mathrm{safe}}$. The operator $\Box_{[0,\tau]}$ requires the predicates throughout the specified horizon. To compare heterogeneous margins, define positive, fixed scales $s_T,s_F,s_d$ and
\[
m_t=\min\!\left\{
\frac{T_{\mathrm{crit}}-T_t}{s_T},
\frac{F_{\max}-\|F_t\|}{s_F},
\frac{d_t-d_{\mathrm{safe}}}{s_d}\right\}.
\]
Here $F_t=F_{\mathrm{contact}}(t)$ and $d_t=d_{\mathrm{vial,obs}}(t)$. Each scale has its corresponding physical unit, making $m_t$ dimensionless. Signed temporal robustness over a fully observed window is its minimum margin~\cite{fainekos2009}. The online monitor uses the available window in $S_t$ and reports both robustness $\rho_t$ and data validity. An incomplete future horizon remains undecided; predicted measurements provide auxiliary information without certifying future safety.

Invalid or stale safety inputs, loss of heartbeat, or $\rho_t<\epsilon$ trigger blocking and a configured safe response, where $\epsilon>0$ is a fixed dimensionless margin. Recovery uses provider-supported compensation or a verified checkpoint; irreversible operations can require safe abort. Monitoring runs independently of policy inference, including waiting periods. Sampling intervals, end-to-end delay, measurement uncertainty, and between-sample violations must be assessed on the target hardware. Schema checks cover discrete parameters; chemical, pressure, and biosafety hazards outside the specification require separate controls and human approval.

We separate safety performance into four quantities: constraint violation rate, interception rate for hazardous actions, false-positive rate for normal actions, and time from failure to a safe state. Accident-free operation must be interpreted together with challenge coverage and effective interception. L3-Bench therefore calls for labeled fault injection in controlled environments, with joint records of gate decisions, device feedback, and operator overrides.

\subsection{Closed-Loop Execution
Algorithm}\label{closed-loop-execution-algorithm}

The following pseudocode specifies runtime synchronization, candidate generation, pre-dispatch gating, and event recording. In Algorithm~\ref{alg:opensdl-runtime}, $\mathcal{G}$ is the task specification, $S_t$ is the synchronized runtime state and history, and $\epsilon$ is the minimum accepted robustness margin. GateAndReserve checks schema, resource availability, and safety conditions, returning approval $ok$ and a logged reason. BuildContext includes confirmed history and available capabilities from $S_t$. Dispatch reconciles pending calls, checks stable idempotency keys, and reserves resources; checkpoints advance only on verified completion. New dispatch requires durable proposal/gate logging; rejected or unlogged candidates release reservations. A preset deadline and retry budget terminate stalled episodes. Recovery depends on provider capabilities.

\begin{algorithm}[H]
\caption{OpenSDL closed-loop execution with Safe-SDL gating}
\label{alg:opensdl-runtime}
\begin{algorithmic}[1]
\Require Task $\mathcal{G}$, safety formula $\varphi_{\mathrm{safe}}$, threshold $\epsilon$
\Ensure $\textsc{Completed}$ or $\textsc{SafeAbort}$
\State $\textsc{InitializeRuntime}()$
\While{$\mathcal{G}$ is not fulfilled}
  \State $(I_t,q_t,M_t,S_t) \gets \textsc{SynchronizeObservation}()$
  \State $(\rho_t,valid) \gets \mathrm{Monitor}(\varphi_{\mathrm{safe}}, S_t)$
  \State $b \gets \textsc{WithinBudget}()$
  \If{$\neg valid \lor \rho_t<\epsilon \lor \neg b$}
    \State $\textsc{BlockPending}();\ \textsc{SafeDegrade}()$
    \State $\textsc{RecoverIfSupported}();\ \textsc{LogDecision}()$
    \State \Return $\textsc{SafeAbort}$
  \EndIf
  \State $e_M \gets \mathrm{TelemetryEncoder}(M_t)$
  \State $X_t \gets \textsc{BuildContext}(I_t,q_t,e_M,S_t)$
  \State $(\hat A_{\mathrm{arm}},\hat{\mathcal{C}}_t) \gets \mathrm{LAM}(X_t,\mathcal{G})$
  \State $(ok,reason) \gets \textsc{GateAndReserve}(\hat A_{\mathrm{arm}},\hat{\mathcal{C}}_t,S_t)$
  \State $logged \gets \textsc{LogProposalAndGate}(ok,reason)$
  \If{$ok \land logged$}
    \State $\textsc{DispatchReady}(\hat A_{\mathrm{arm}},\hat{\mathcal{C}}_t)$
  \Else
    \State $\textsc{AbstainOrRequestReview}(reason)$
  \EndIf
  \State $\textsc{RecordReturnsAndCheckpoint}()$
\EndWhile
\State \Return $\textsc{Completed}$
\end{algorithmic}
\end{algorithm}

\subsection{Reproducible Research
Loop}\label{reproducible-research-loop}

The research workflow uses the same schema and event ledger as the runtime. Here HIL (hardware-in-the-loop) means that real controllers or devices are connected to a simulated environment, whereas a digital twin is a software model used for replay or simulation. The research lane covers trajectories, schemas, models, and evaluation, while the runtime lane covers observations, device routing, and replayable records. Model inference and L3-Bench results connect the two lanes through the same manifest and event ledger. In Algorithm~\ref{alg:opensdl-research}, $\mathcal{D}$ supplies held-out evaluation cases, $\mathcal{D}'$ their schema-normalized specifications, $\mathcal{W}$ the LAM checkpoint, $\mathcal{E}$ is the simulator or HIL context, $\pi$ is the loaded policy, $\mathcal{E}'$ is the seeded digital twin, $\mathcal{Z}_{\mathrm{sim}}$ and $\mathcal{Z}_{\mathrm{hil}}$ are simulation and optional HIL rollouts, and $\mathcal{R}$ is the exported benchmark report with its configuration, seed, asset, and log manifest. Rollouts cover the four registered evaluation settings defined in Section~\ref{metrics}, with the concurrency ablation using matched task instances and seeds.

\begin{algorithm}[H]
\caption{OpenSDL reproducible research loop}
\label{alg:opensdl-research}
\begin{algorithmic}[1]
\Require Held-out cases $\mathcal{D}$, LAM checkpoint $\mathcal{W}$, simulator or HIL context $\mathcal{E}$
\Ensure Benchmark report $\mathcal{R}$ and replay artifacts
\State $\mathcal{D}' \gets \textsc{NormalizeSDLTraj}(\mathcal{D})$
\State $\textsc{ValidateSchema}(\mathcal{D}')$
\State $\pi \gets \textsc{LoadLAM}(\mathcal{W},\mathrm{measurement\ context})$
\State $\mathcal{E}' \gets \textsc{SeededTwin}(\mathcal{E},\mathrm{Clean\mbox{-}3D})$
\State $\mathcal{Z}_{\mathrm{sim}} \gets \textsc{Rollout}(\pi,\mathcal{E}',\mathcal{D}',\mathrm{settings})$
\State $\mathcal{Z}_{\mathrm{hil}} \gets \textsc{OptionalHIL}(\pi,\mathcal{E},\mathcal{D}')$
\State $\mathcal{R} \gets \textsc{L3Bench}(\mathcal{Z}_{\mathrm{sim}}\cup\mathcal{Z}_{\mathrm{hil}})$
\State \Return $\textsc{Export}(\mathcal{R},\mathrm{config,seed,assets,logs})$
\end{algorithmic}
\end{algorithm}

\subsection{Artifact Manifest and Release Tiers}\label{artifact-manifest-and-release-tiers}

The proposed run manifest binds an episode ID to code and schema versions, model hash, provider inventory, calibration, digital-twin assets, random seeds, fault configuration, and event-log indices. Clean-3D names the planned versioned scene/geometry asset set. Missing required fields are reported as reproducibility failures. Tier-1 releases contain schemas, simulators, and sanitized trajectories; Tier-2 adds HIL configurations and logs; Tier-3 adds permitted physical-device providers, calibration, and raw records. Restricted protocols require simulated substitutes with role descriptors and fault scripts, with each result identifying its tier and limitations.

\begin{figure*}[t]
\centering
\includegraphics[width=\textwidth]{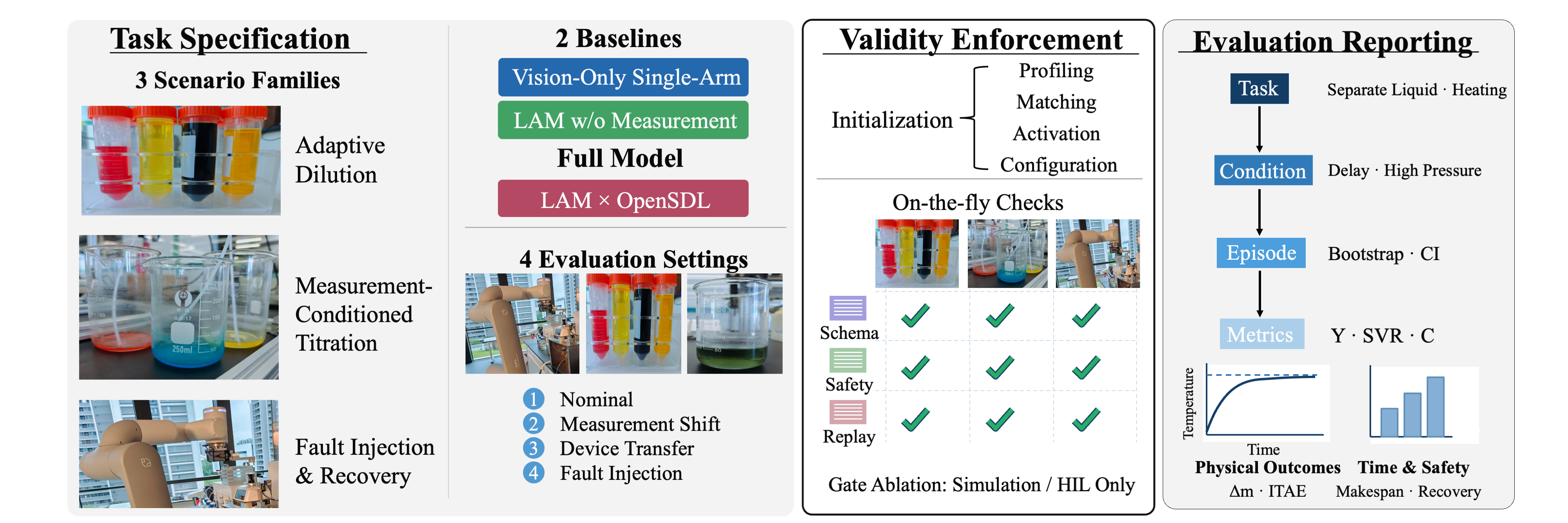}
\caption{Overview of the L3-Bench evaluation protocol. Left: adaptive dilution, measurement-conditioned titration, and fault injection with recovery define three scenario families. Two baselines---a vision-only single-arm policy and LAM without measurement tokens---are compared with the full LAM $\times$ OpenSDL framework under nominal, measurement-shift, device-transfer, and fault-injection conditions. Center: initialization comprises profiling, matching, activation, and configuration, followed by on-the-fly schema, safety, and replay-readiness checks of manifest and ledger completeness. The safety-gate ablation is restricted to simulation or risk-assessed hardware-in-the-loop (HIL) evaluation. Right: liquid separation/heating and delay/high pressure illustrate task and condition metadata; the evaluated scenario families are defined on the left. Complete episodes are scored, with batch-level clustering for bootstrap confidence intervals. Primary endpoints are normalized physical yield $Y$, temporal safety violation rate SVR, and operational time cost $C$; diagnostic measures include mass-transfer error $\Delta m$, temperature-control error ITAE, makespan, and recovery. Photographs illustrate the laboratory scenarios; check marks and miniature plots indicate protocol requirements and metric types.}
\label{fig:l3bench-overview}
\end{figure*}

\begin{table*}[t]
\centering
\small
\setlength{\tabcolsep}{4pt}
\caption{Pre-specified evaluation matrix for L3-Bench.}
\label{tab:evaluation-matrix}
\begin{tabular}{p{0.19\textwidth}p{0.27\textwidth}p{0.28\textwidth}p{0.18\textwidth}}
\toprule
\textbf{Evaluation setting} & \textbf{Controlled intervention} & \textbf{Primary question} & \textbf{Required report} \\
\midrule
Nominal & Same devices, protocol, and measurement range & Can the full loop execute the task? & $Y$, $C$, completion rate \\
Measurement shift & Noise, latency, or target-range change & Do measurement tokens improve adaptation (\ref{rq:measurement})? & $\Delta m$, $CV$, calibration error \\
Fault injection & Offset, slip, timeout, over-temperature & Does gating contain and recover from faults (\ref{rq:safety})? & SVR, interception, recovery time \\
Device transfer & New provider with the same role descriptor & Are action semantics portable (\ref{rq:portability})? & Schema failures, transfer gap \\
\bottomrule
\end{tabular}
\end{table*}

\section{L3-Bench and Evaluation
Plan}\label{l3-bench-and-evaluation-plan}

\subsection{Metrics and Evaluation Protocol}\label{metrics}\label{benchmark-unit-and-reporting-rules}\label{evaluation-matrix}

L3-Bench (Level-3 laboratory benchmark) complements binary task success with verifiable physical quantities. Figure~\ref{fig:l3bench-overview} summarizes the scenario families, comparative conditions, validity checks, and reporting hierarchy. Mass-transfer error is $\Delta m=|m_{\text{target}}-m_{\text{measured}}|$, the absolute deviation between the protocol target mass and the measured delivered mass. Dispensing uniformity is $CV=\sigma_V/\mu_V$, with sample standard deviation $\sigma_V$ across wells and positive mean volume $\mu_V$; report the mean and target-volume error alongside $CV$, since uniformly incorrect dispensing can have low variation. For temperature control, overshoot is $\max_{t\in[0,t_f]}\max(0,T(t)-T_{\text{set}})$, settling time is the first sustained entry into a fixed tolerance band over a prescribed observation window, and
\[
ITAE=\int_0^{t_f}t|T(t)-T_{\text{set}}|\,dt
\]
weights temperature error by elapsed time over the observation duration $t_f$. The set point is the requested thermal condition. Overshoot measures excursion above it, settling time measures response speed, and ITAE has units of temperature times time squared. Runs that never settle are reported as censored; safety margin is evaluated against the separate critical limit. Temporal safety violation rate (SVR) is the fraction of evaluated operations with at least one constraint violation, using protocol-level operation boundaries shared across methods; reports also include episode-level violation incidence and monitored violation duration, including waits. Replay compares event IDs, causal order, and recomputed metrics against fixed tolerances. Independent reruns assess paired metric differences against pre-specified equivalence margins; equivalence requires the confidence interval to lie entirely within those margins. Thresholds must be pre-registered according to device precision and the experimental protocol before results are observed.

The basic statistical unit in L3-Bench is a complete episode, from task submission to completion, rollback, or safe termination. A successful grasp or individual API call is an intermediate metric; the episode-level outcome remains the evaluation unit. For episode $k$, normalized yield and time cost are defined as
\[
\begin{aligned}
Y_k&=\frac{y_k-y_{\min}}{y_{\max}-y_{\min}},\\
C_k&=t_k+\eta_1 n_{\mathrm{retry},k}+\eta_2 n_{\mathrm{human},k},
\end{aligned}
\]
Here $y_k$ is a protocol-defined, higher-is-better physical outcome for episode $k$, while $y_{\min}$ and $y_{\max}$ are fixed lower and upper reference values with $y_{\max}>y_{\min}$. Values outside $[0,1]$ are retained and flagged; normalization is protocol-specific. The makespan $t_k$ is elapsed time from task submission to completion or safe termination; $n_{\mathrm{retry},k}$ and $n_{\mathrm{human},k}$ count device retries and human interventions. The coefficients $\eta_1$ and $\eta_2$ convert those interventions into the same cost units as time and must be fixed before experimentation and reported. These penalties represent additional intervention cost, including any burden beyond elapsed time. Failed and aborted episodes retain their measured outcomes and labels; missing outcomes follow a pre-specified reporting rule. Cost comparisons separate completed and aborted runs.

Each configuration must report at least the episode count, mean and median, interquartile or confidence intervals, a breakdown of failure causes, and safety-kernel false positives and missed detections. Continuous and binary metrics are aggregated separately. If device incompatibility prevents a scenario from being executed, the result is marked not applicable with a retained explanation.

Table~\ref{tab:evaluation-matrix} maps interventions to the research questions defined in Section~\ref{design-requirements}. Nominal conditions test basic closed-loop execution. Measurement shift changes noise, sampling frequency, or target ranges to test \ref{rq:measurement}. Fault injection triggers displacement, slips, timeouts, or overheating to test \ref{rq:safety}. Device transfer preserves role capabilities while substituting the provider to test \ref{rq:portability}. The concurrency ablation evaluates \ref{rq:coordination}, and fixed-manifest replay checks the reproducibility component of \ref{rq:portability}. All comparisons use fixed task scripts, thresholds, and episode termination rules.

\subsection{Scenarios and Comparative Evaluation}\label{scenarios-and-baselines}\label{ablations-and-statistical-protocol}

Evaluation should cover at least three scenario families: \textbf{adaptive high-throughput dilution}, where a pipetting station serving 96- or 384-well plates and robot arm coordinate gradient dispensing and transfers between workstations; \textbf{measurement-conditioned titration}, where feed rates are adjusted using pH, mass, color, or temperature; and \textbf{fault injection and recovery}, where displaced consumables, grasp slips, abnormal cap resistance, and communication timeouts are simulated to compare safe stopping, retry, and rollback.

The two primary baselines are a vision-only single-arm policy and LAM without measurement tokens, both compared with the full LAM $\times$ OpenSDL framework. The measurement-token ablation removes measurement tokens from policy context while retaining runtime logging and safety checks. It also masks measurement values and derived summaries in event history, holding architecture, training budget, devices, and task instances fixed. The single-arm baseline tests overall system capability with different device access and concurrency; component attribution uses matched ablations. Throughput reports must also state batch duration, the number of parallel devices, and failed-attempt and retry costs, avoiding direct ratios between unequal workloads.

Component ablations isolate the effects of removing measurement tokens, disabling concurrency between continuous and discrete heads, disabling runtime-ledger integration while retaining an independent evaluation logger, and an offline control with the Safe-SDL gate disabled. To address \ref{rq:coordination}, the concurrency ablation varies overlapping action dependencies and reports makespan, device utilization, and retries. The gate-disabled condition is restricted to simulation or a risk-assessed HIL environment to estimate the safety kernel's effect on normal throughput. Physical interlocks remain active in HIL. Each ablation changes one component and fixes model capacity, task instances, devices, seeds, and other execution rules. Identical control commands are used when measuring logging overhead.

Statistical analysis preserves paired comparisons and resamples independent experimental batches when their episodes share conditions. Reports stratify by task, device, and fault type; a pre-specified hierarchical model can account for these dependencies. Reports include effect sizes, uncertainty intervals, and worst-case quantiles, since average yield can hide rare but severe safety failures. The primary endpoints ($Y$, SVR, and $C$) are specified in advance; other metrics serve as secondary endpoints or diagnostics. Any changes to thresholds, exclusions of failed episodes, or revisions to termination rules after inspecting the test set must be documented in the version history.

\subsection{Current Status and Reproducibility Checklist}\label{current-status-and-reproducibility}\label{reproducibility-checklist}\label{current-status-and-pending-experiments}

For \ref{rq:portability}, the reproducibility checklist in Appendix~\ref{app:release-checklist} requires versioned configurations, calibrations, seeds, complete event records, and result-generation scripts. A release must regenerate summaries from manifests and run schema, provider, and replay checks. Dry-run or simulated replay results identify the hardware conditions they leave untested.

Table~\ref{tab:status} summarizes implementation status, available evidence, and pending experiments for each component.

\begin{table}[H]
\centering
\footnotesize
\setlength{\tabcolsep}{2pt}
\caption{Implementation status and evidence planned for the arXiv release.}
\label{tab:status}
\begin{tabular}{@{}>{\raggedright\arraybackslash}p{0.30\linewidth}>{\raggedright\arraybackslash}p{\dimexpr0.70\linewidth-2\tabcolsep\relax}@{}}
\toprule
\textbf{Component} & \textbf{Status and evidence} \\
\midrule
OpenSDL 1.0 mother/\allowbreak{}edge/\allowbreak{}transport & Control engine, adapters, transports, and SQLite logging implemented; end-to-end replay validation pending. \\
Role/\allowbreak{}DeviceProvider and SDL-Traj & Proposed research contracts; full schema and compatibility validation pending. \\
LAM hybrid action heads & Proposed architecture and training objectives; trained checkpoints and ablations pending. \\
Clean-3D assets and digital twin & Asset specification; released manifests and validated scene inventory required. \\
L3-Bench & Metrics, scenarios, and evaluation protocol defined; full results pending. \\
Hardware transfer and safety statistics & Hardware-transfer experiments and safety statistics remain pending. \\
\bottomrule
\end{tabular}
\end{table}

\section{Discussion and Limitations}\label{discussion-and-limitations}\label{discussion-open-science-and-limitations}

The central implication of this work is that an autonomous laboratory interface must jointly express heterogeneous instrument calls, continuous robot trajectories, waiting semantics, and safety constraints, while preserving a shared schema across simulation, HIL, and hardware. LAM $\times$ OpenSDL therefore defines falsifiable engineering hypotheses: measurement tokens should improve adaptation under measurement shift; hybrid action graphs should reduce makespan under concurrency stress; Safe-SDL should turn faults into interpretable blocks, degradation, or rollback; and role-preserving provider substitution and fixed-manifest replay should preserve action semantics, key state transitions, and metrics. End-to-end training, cross-device transfer, fault injection, and episode-level statistics remain necessary to establish these benefits. When outcomes fall short, failures should be attributed to representation, scheduling, providers, devices, or statistical protocols.

The scope of these claims is limited. The target runtime combines schema validation, least-privilege providers, heartbeats, STL monitoring, and human oversight. Current SQLite logging provides append-only application behavior; integrity under privileged modification requires additional access controls and tamper-evident storage. Complete sensor failures, unknown chemical reactions, and hazards outside the formalized model require safe termination; safety certification remains outside the scope of the architecture. Public releases should include role schemas, simulated providers, example configurations, Clean-3D asset manifests, generation scripts, and fault injectors, using hashes, interfaces, and sanitized substitutes for unauthorized proprietary protocols, sensitive processes, and industrial models. This manuscript remains an architecture and evaluation protocol. The external validity of L3/L4 levels for chemical and biological tasks, protocol-dependent physical yield, the coverage of STL constraints, and reduced concurrency under shared workspaces or irreversible reactions require further experiments and human oversight.

\section{Conclusion}\label{conclusion}

This work proposes LAM $\times$ OpenSDL to connect continuous robot control, typed instrument calls, and measurement feedback in heterogeneous SDLs. The proposed LAM interface conditions motion and instrument actions on shared observations and typed measurement tokens carrying units, timestamps, and quality flags. Its action graphs make dependencies, resource use, waiting, and completion conditions explicit.

The initial OpenSDL control substrate provides device routing, protocol adapters, transports, and SQLite logging. The proposed research contracts add role-level portability, causal records, and comparison across simulation, HIL, and hardware. Safe-SDL specifies pre-dispatch checks and continuous monitoring with conditional recovery; L3-Bench defines episode-level physical outcomes, timing, safety, and reproducibility reports. These interfaces connect performance hypotheses to the command and measurement records needed to test them.

The present work establishes a research foundation comprising the OpenSDL control substrate, LAM interface designs, and the L3-Bench evaluation protocol; throughput, adaptability, and safety benefits still require systematic validation. Future work will address model training, measurement-conditioning ablations, hybrid concurrent scheduling, cross-device transfer, and controlled fault injection. Comparisons among vision-only policies, LAM without measurement tokens, and the complete framework will use fixed configurations, seeds, and calibration conditions, while reporting failures, human interventions, and recovery costs. By combining reusable interfaces with traceable evaluation, LAM $\times$ OpenSDL provides an implementable and testable approach to measurement-driven, heterogeneous, and reproducible laboratory control.

\section*{Acknowledgments}
\label{acknowledgments}

This work was supported by the National Natural Science Foundation of China (Grant Nos. 92570205 and 92461313) and the Chuangqi Program of Shanghai Innovation Institute, Shanghai Municipal Education Commission.

\fontsize{10}{11}\selectfont
\flushbottom

\par\penalty-10000\relax
\clearpage
\raggedbottom

\appendix
\noindent\begin{minipage}[b][\dimexpr\textheight-1.93pt\relax][s]{\columnwidth}
\fontsize{8.5}{10.25}\selectfont
\section{Illustrative Role Descriptor}\label{app:role-descriptor}

The following minimal role descriptor is consistent with the proposed interface. It is a semantic example spanning devices; concrete providers may compile the same calls into serial registers, network APIs, or HIL simulators.

\begin{verbatim}
role: heater
schema_version: sdl-role/v1
operations:
  - name: Set
    parameters:
      target_temperature:
        unit: degC
        min: 4
        max: 150
      hold_time:
        unit: s
        min: 0
    completion: stable_for
    idempotency: replace
  - name: Read
    returns: [temperature, unit, quality,
              sample_time, calibration_version]
resources: [heater_zone_0]
rollback: set_safe_temperature
provider: simulator.heater.v1
\end{verbatim}

The descriptor states capabilities and safety ranges independently of provider internals. A tester can use the same descriptor to generate schema-validity, out-of-range parameter, timeout, and rollback tests, making device substitution a provider-compatibility question that can be checked automatically.

\section{Failure Taxonomy and Labeling}\label{app:failure-taxonomy}

Failed or aborted episodes receive one primary execution-failure label and may carry secondary labels. REPLAY separately flags reconstruction failure, including for a physically completed episode. The labels are defined below.

\begin{table}[H]
\centering
\small
\setlength{\tabcolsep}{2pt}
\caption{Failure labels used for episode-level analysis.}
\label{tab:failure-taxonomy}
\begin{tabular}{p{0.22\columnwidth}p{0.66\columnwidth}}
\toprule
\textbf{Label} & \textbf{Operational definition} \\
\midrule
MODEL & LAM proposes an incorrect role, trajectory, or parameter despite valid inputs. \\
SCHEMA & Type, unit, range, or required-field validation rejects the proposal. \\
RESOURCE & A device, workspace, or barrier dependency cannot be acquired. \\
DEVICE & Provider returns a hardware, calibration, or communication failure. \\
SAFETY & STL, heartbeat, or data-quality gate blocks or aborts execution. \\
REPLAY & The episode cannot be reconstructed from its manifest and event ledger. \\
\bottomrule
\end{tabular}
\end{table}

For multiple execution-failure labels, the primary label identifies the first causal event that prevents normal completion. Subsequent faults receive secondary labels, and compensating actions remain in the event ledger. This rule avoids misreporting safety interception as device failure and prevents rollback cost from being hidden within model retry counts.

\par\end{minipage}
\newpage
\noindent\begin{minipage}[b][\dimexpr\textheight-1.93pt\relax][s]{\columnwidth}
\fontsize{8.5}{10.25}\selectfont
\setlength{\parskip}{0pt plus 1fil}
\section{Reproducible Release Checklist}\label{app:release-checklist}

A release candidate is reproducible only when the software, physical configuration, and episode evidence can be reconstructed together. The following checklist turns the protocol into a concrete release gate:

\begin{itemize}
\setlength{\itemsep}{1pt}
\item \textbf{Code and schemas:} record repository commits, schema versions, interface contracts, and model checkpoint hashes.
\item \textbf{Assets and devices:} record Clean-3D digital-twin asset versions, provider versions, device inventories, calibration files, units, and clock offsets.
\item \textbf{Configuration:} record task scripts, safety formulas, thresholds, random seeds, sampling temperature, and inference batch size.
\item \textbf{Episode evidence:} retain synchronized observations, proposed and dispatched actions, device returns, event-ledger indices, operator overrides, and termination labels.
\item \textbf{Replay checks:} run schema validation, provider compatibility, dry-run, sim-replay, and hardware-replay tests when the required devices are available.
\item \textbf{Result generation:} regenerate tables and figures from manifests with versioned scripts, and report unavailable hardware or sanitized proprietary providers explicitly.
\end{itemize}

\section*{Declarations}\label{declarations}

\textbf{Data Availability.} This is a system design and engineering report. It uses no human or animal subject data and reports no new biomedical measurements. The planned public release comprises a versioned code repository with releasable schemas, example configurations, figures, and sanitized trajectories. Providers, calibration files, and raw device bytes subject to commercial restrictions are represented by interfaces, hashes, or simulated substitutes. Any episode used to generate a reported number must include a manifest, an event-log index, and data-processing scripts.

\textbf{Ethics Statement.} The current work involves no human participants, animal experiments, or personal data. Physical deployment must comply with institutional rules for chemicals, biosafety, pressure vessels, and device interlocks. Safe-SDL provides software-level gating; institutional approval and on-site training remain required.

\textbf{Author Contributions.} The author team jointly developed the problem formulation, system architecture, OpenSDL interface design, evaluation protocol, and manuscript.

\textbf{Competing Interests.} The authors declare no competing interests directly related to the design objectives of this work.

\textbf{Funding.} Funding sources and grant numbers are provided in the Acknowledgments.

\textbf{AI Assistance.} AI-assisted tools were used for language editing, format conversion, and preparation of selected figures. The authors are responsible for checking the technical content, references, code, figures, tables, and final submission, and take responsibility for all claims in the paper.

\par\end{minipage}

\end{document}